\documentclass[10pt,twocolumn,letterpaper]{article}

\usepackage[pagenumbers]{wacv} 

\usepackage{graphicx}
\usepackage{amsmath}
\usepackage{amssymb}
\usepackage{booktabs}

\usepackage[accsupp]{axessibility}  

\usepackage[pagebackref,breaklinks,colorlinks]{hyperref}

\usepackage{breqn}
\usepackage{amsmath}

\usepackage{algorithm}
\usepackage{algpseudocode}
\usepackage{comment}
\usepackage{color}
\usepackage{float}
\usepackage[accsupp]{axessibility}  

\usepackage{float}

\usepackage{blindtext}

\usepackage[capitalize]{cleveref}
\crefname{section}{Sec.}{Secs.}
\Crefname{section}{Section}{Sections}
\Crefname{table}{Table}{Tables}
\crefname{table}{Tab.}{Tabs.}
\usepackage{enumitem}

\def\wacvPaperID{813} 
\def\confName{WACV}
\def\confYear{2025}

\begin{document}

\title{Localized Gaussian Splatting Editing with Contextual Awareness}

\author{Hanyuan Xiao\textsuperscript{1,2}\quad Yingshu Chen\textsuperscript{3}\quad Huajian Huang\textsuperscript{3}\\Haolin Xiong\textsuperscript{4}\quad Jing Yang\textsuperscript{1,2}\quad Pratusha Prasad\textsuperscript{1,2}\quad Yajie Zhao\textsuperscript{1,2,\thanks{Corresponding Author}}\\\\
\textsuperscript{1}University of Southern California\quad \textsuperscript{2}Institute for Creative Technologies\\ \textsuperscript{3}HKUST\quad \textsuperscript{4}University of California, Los Angeles\\
{\tt\small \textsuperscript{1,2}\{hxiao, jyang, bprasad, zhao\}@ict.usc.edu\quad \textsuperscript{3}\{ychengw, hhuangbg\}@connect.ust.hk\quad \textsuperscript{4}\{xiongh\}@ucla.edu}
}

\vspace{-1cm}
\makeatletter
\let\@oldmaketitle\@maketitle
\renewcommand{\@maketitle}{\@oldmaketitle
  \centering
  \includegraphics[width=0.9\linewidth]
    {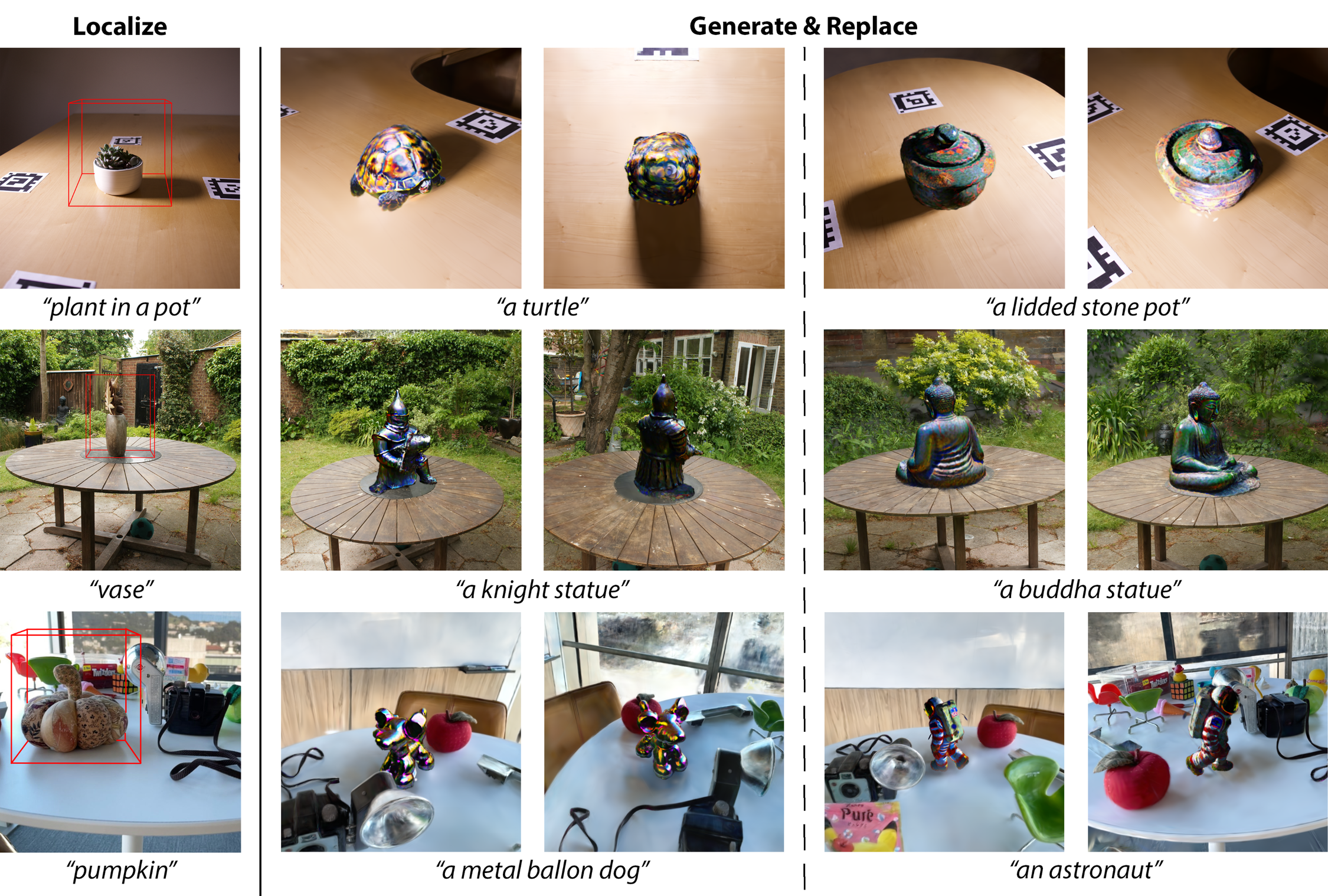}
  \captionof{figure}{We present a novel pipeline for text-guided localized Gaussian splatting scene editing. It replaces and edits regions of interest based on user text inputs, achieving realistic and high quality visual results that naturally match the original context, including illumination and occlusions. Compared to previous methods, we are the first to enable object replacement with consistent illumination to the surroundings.}} 
\makeatother

\maketitle



\begin{abstract}
    Recent advancements in text-guided 3D object generation using diffusion priors struggle with illumination inconsistencies when applied to scene editing tasks like object replacement or insertion. To address this, we propose an illumination-aware 3D scene editing pipeline for 3D Gaussian Splatting (3DGS). Our method leverages state-of-the-art 2D diffusion inpainting~\cite{zhang2023adding} to handle global illumination context effectively. Specifically, we identify representative anchor views that capture scene-wide illumination, inpaint them using 2D diffusion models, and integrate the results into a coarse-to-fine 3DGS optimization process. In the fine step, we introduce Depth-guided Inpainting Score Distillation Sampling (DI-SDS) to refine geometry and texture details, capitalizing on the diversity of 2D priors. Our approach achieves locally precise edits with globally consistent illumination, demonstrating robustness in real scenes with highlights and shadows. Comparisons show superior results over state-of-the-art text-to-3D editing methods. Project page: \href{https://corneliushsiao.github.io/GSLE.html}{https://corneliushsiao.github.io/GSLE.html}.
\end{abstract}


\vspace{-.3cm}
\section{Introduction} \label{sec:introduction}

Recent advancements in text-guided 3D object generation have shown remarkable success by leveraging the power of diffusion models \cite{poole2023dreamfusion,wang2023prolificdreamer, liu2023zero1to3, wang2023score, lin2023magic3d,tang2024dreamgaussian}. However, little attention has been paid to text-guided localized scene editing task. Naive insertion of generated 3D objects leads to mismatch of illumination appeared in the scene. Global editing methods that do not include localization module~\cite{instructnerf2023,wang2023inpaintnerf360,dong2023vicanerf} fail by adding color clue to the whole scene or changing undesired regions. Existing solutions for text-guided 3D editing \cite{bao2023sine, song2023blending, zhuang2023dreameditor, fang2023gaussianeditor} can only achieve minor alternation in geometry or texture of existing objects. Tasks such as context-aware object insertion and object replacement with pronounced change are still underexplored. 


In this work, 3D Gaussian Splatting (3DGS)~\cite{kerbl3Dgaussians} represents both scene and generated object, because 1) it supports efficient optimization in 3D lifting task, and 2) provides explicit representation that enables editing independent of background, in contrast to NeRF representation~\cite{mildenhall2021nerf}.
To achieve holistically photorealistic 3D editing, it is necessary to take global information such as illumination, occlusions into account. 
Conventional approaches involving inverse rendering, material generation and physically-based neural rendering are time consuming and unstable~\cite{liang2023gs,chen2023text2tex}.
Instead, we observed large-scale diffusion models conditioned on background and geometry information (e.g., depth or camera pose) excel at the inpainting task, particularly in regards to maintaining illumination matching and seamlessly fusing the generated foreground and background with the original image.
Building upon this observation, we introduce a 3D scene editing pipeline that considers scene illumination. By advancing algorithm of 3D lifting and score distillation sampling (SDS)~\cite{poole2023dreamfusion}, 
our pipeline accomplishes challenging object insertion and replacement by incorporating prior knowledge from both a depth-guided inpainting diffusion model~\cite{zhang2023adding} and a 3D-aware diffusion model~\cite{liu2023zero1to3}.

For robust and harmonious object synthesis, we introduce Anchor View Proposal (AVP) algorithm, which selects a view that best represents global illumination with the most contrast. 
This picked view will be inpainted given the input text prompt, working as the initial guidance for the following two-step contextual 3D lifting. By analysis, this inpainted guidance offers heuristic information for illumination-aware object synthesis.
Among the two steps, we first conduct a coarse image-to-3D generation and then an illumination-aware texture enhancement optimization. 
In the coarse step, we explore score distillation of the 3D-aware diffusion prior from the multi-view diffusion model~\cite{liu2023zero1to3} given the foreground object in the inpainted anchor view. This diffusion prior comprehends essential illumination information from anchor view and guarantees multi-view consistency in 3D object synthesis. 
In the enhancement step, we propose a new score distillation scheme (DI-SDS) that leverages depth-guided inpainting diffusion prior with background context. This strategy yields multi-view-consistent and contextually illumination-aware object refinement.

To summarize, our main contributions include:
\begin{enumerate}[leftmargin=*]
    \item We present a comprehensive pipeline to generate objects based on text inputs that seamlessly integrate into a 3D Gaussian Splatting scene. In particular, our method enables objects generation that automatically matches global illumination. 
    \item We introduce Anchor View Proposal algorithm for automatic selection of a representative view that best represents illumination at a target region within a complex scene. The anchor view selection provides features of scene illumination for harmonious object synthesis.
    \item We propose Depth-guided Inpainting Score Distillation Sampling, which embeds both condition of geometry and contextual illumination for object generation and texture enhancement.
\end{enumerate}
\begin{figure*}[ht]
  \centering
  \includegraphics[width=\linewidth]{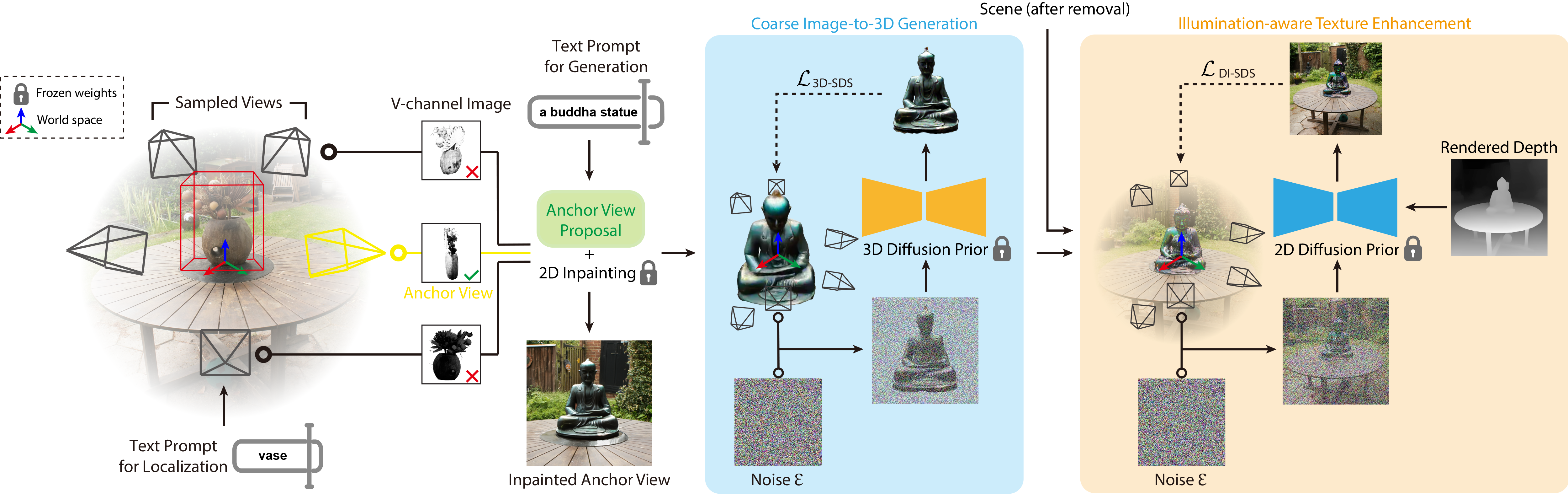}
  \vspace{-0.3cm}
  \caption{\textbf{Overview of Pipeline}. 
  We propose a pipeline for localized scene editing, the new objects match the global illumination in the original scene. Our method uses text prompts to guide localization and generation, enabling us to synthesize and insert an object that is aware of the surrounding lighting. We introduce Anchor View Proposal to find a representative view of the illumination, which enables a two-step 3D lifting pipeline. Specifically, we propose illumination-aware Texture Enhancement with Depth-guided Inpainting Score Distillation Sampling (DI-SDS), which leverages contextual inpainting capability of large diffusion model. This enhancement step not only improves delicate texture details and motifs but also preserves global lighting conditions during optimization.}
  \label{fig:pipeline}
\end{figure*}

\section{Related Works} \label{sec:related-works}


Recent breakthroughs in large-scale vision-language models (VLMs)~\cite{radford2021learning,li2022blip} and text-guided generative models featuring Stable Diffusion \cite{rombach2022high} have boosted the research field in 2D-to-3D generation and editing, leveraging these powerful priors. We review recent literature on image- and text-guided 3D generation and 3D scene editing, focusing on scenes using radiance field representation. 
\subsection{3D Object Generation} Early exploration of text-to-3D utilizes pre-trained VLMs such as CLIP \cite{radford2021learning} for zero-shot 3D model generation \cite{sanghi2022clip,jain2022zero,khalid2022clipmesh}, but with simple shape and texture styles. 
Subsequently, generative diffusion models equipped with advanced neural field representations \cite{xie2022neural} have demonstrated remarkable capabilities in high-fidelity 3D generation tasks via 2D-to-3D lifting, encompassing text-to-3D~\cite{poole2023dreamfusion, wang2023prolificdreamer, liu2023zero1to3, zhao2023michelangelo, lorraine2023att3d, hong2024debiasing, wang2023score, lin2023magic3d,tang2024dreamgaussian} and image-to-3D domains~\cite{liu2024syncdreamer, zhao2023michelangelo, qian2023magic123, liu2023one2345,tang2024dreamgaussian}. Notably, DreamFusion \cite{poole2023dreamfusion} and SJC \cite{wang2023score} introduced the Score Distillation Sampling (SDS) optimization scheme by distilling 2D diffusion knowledge, yielding encouraging text-to-3D results. 
The follow-up works further extended SDS to resolve issues such as low resolution and over-saturation. For example, Magic3D~\cite{lin2023magic3d} employs a coarse-to-fine strategy using image and latent diffusion priors for high-quality generation, ProlificDreamer~\cite{wang2023prolificdreamer} addresses issues of over-saturation, over-smoothing, and low diversity through variational score distillation (VSD), and D-SDS~\cite{hong2024debiasing} enhances SDS by incorporating score debiasing and prompt debiasing mechanisms. 
In parallel, with a single image input, Zero123~\cite{liu2023zero1to3} excels in generating realistic novel perspectives via a view-dependent diffusion model, while SyncDreamer~\cite{liu2024syncdreamer} synthesizes view-consistent novel perspectives without relying on neural fields through an intermediate synchronization statistics approach. With recent advanced 3D Gaussian Splatting (3DGS) \cite{kerbl3Dgaussians} scene representation featuring swift training and rendering speed, \cite{tang2024dreamgaussian, chen2023text} increase generation speed and achieve text-to-3D generation with delicate texture details. 
These works primarily focus on single-object generation neglecting background environment interactions, such as ambient illumination and atmosphere. In contrast, our approach incorporates background effect into the 3D generation process with holistic contextual harmony between generation and background scene. 


\subsection{3D Scene Editing} 
The rising popularity of implicit 3D representations, particularly radiance fields \cite{mildenhall2021nerf, kerbl3Dgaussians}, has increased interest in 3D scene geometry and texture editing \cite{fan2022unified, nguyen2022snerf,liu2023stylerf, wang2021clip, liu2021editing, bao2023sine, zhuang2023dreameditor}. Advances in this field utilizes pre-trained vision, language and generative models \cite{simonyan2014very, caron2021emerging, radford2021learning, li2022blip, rombach2022high}, and achieves either direct modifications of entire scene ~\cite{wang2023nerf, instructnerf2023, dong2023vicanerf} or intricate manipulations and compositions of scene elements~\cite{bao2023sine, kim20233daware, gordon2023blended, wang2023inpaintnerf360, michel2023object, zhuang2023dreameditor}.
Given a textual editing instruction, NeRF-Art~\cite{wang2023nerf} streamlines the modification of simple scenes by combining a foundational NeRF model with a style-oriented NeRF steered by CLIP priors. 
To progressively edit a NeRF scene, Instruct-NeRF2NeRF~\cite{instructnerf2023} alternatively updates training multi-views by Instruct-Pix2Pix~\cite{Brooks_2023_CVPR} and then finetunes NeRF scene with updated training views. Later, to address the issue of multi-view inconsistent editing, ViCA-NeRF~\cite{dong2023vicanerf} enhances this process by integrating multi-view awareness with a blending mechanism.
For more intricate scene alterations, SINE~\cite{bao2023sine} advances image-guided 3D semantic editing by using an editing field informed by geometry and semantic vision priors \cite{deng2021deformed, caron2021emerging} to infuse detailed geometric and textural shifts. 
Meanwhile, some 3D editing works \cite{song2023blending, michel2023object, zhuang2023dreameditor, shum2024language, shahbazi2024inserf} focus on precise local changes with 3D segmentation and semantic editing. Within a target region in a NeRF scene, \cite{song2023blending} achieves both geometry and texture alterations, and \cite{shum2024language, shahbazi2024inserf} support object addition. DreamEditor~\cite{zhuang2023dreameditor} polishes local editing with a mesh-based field distilling from NeRF that can edit specific explicit segments within the 3D landscape. Concurrent works \cite{fang2023gaussianeditor, chen2023gaussianeditor} employ 3DGS for swift text-guided scene editing. Compared to these editing works, we resolve the problem of object generation and replacement with dramatic geometry changes and background alignment.

\vspace{-.3cm}

\section{Preliminaries} \label{sec:preliminary}


\subsection{Score Distillation Sampling} \label{sec:prel-sds}
Text-guided 3D generation has attracted increasing research interest and demonstrated significant progress by optimizing a 3D representation $\theta$ using a 2D pre-trained image diffusion prior $\epsilon_\phi$ based on Score Distillation Sampling (SDS), as originally proposed in DreamFusion \cite{poole2023dreamfusion}. The diffusion model $\phi$ is pre-trained to predict sampled noise $\epsilon_\phi (x_t;t,y)$ at timestep $t$ of the noisy image $x_t$, conditioned on text embeddings $y$. By rendering a random view by a differentiable renderer $g(\cdot)$, the aim is to minimize the discrepancy between the rendered image $\mathbf{x}=g(\theta)$ and the diffusion model distribution through the SDS loss. The scene parameterized by $\theta$ can be updated by computing the gradient derived from SDS loss:
\begin{equation}\label{eqn:sds}
    \nabla_\theta \mathcal{L}_{SDS}(\phi, \mathbf{x})=\mathbb{E}_{t,\epsilon}[w(t)(\epsilon_\phi (x_t;t,y)-\epsilon)\frac{\partial \mathbf{x}}{\partial \theta}],
\end{equation}
where $w(t)$ is a weighting function, $\epsilon$ is a Gaussian noise. Our method adapted SDS loss to incorporate the 3D-aware diffusion prior and a customized inpainting diffusion prior, compatible with 3D Gaussian Splatting representations. 


\subsection{3D Gaussian Splatting} \label{sec:prel-3dgs}

3D Gaussian Splatting (3DGS) \cite{kerbl3Dgaussians} represents the scene with a set of anisotropic 3D Gaussians $\mathbf{G}(x)$ which can be efficiently projected into 2D Gaussians $\mathbf{G'}(x')$ for rendering. The differential rasterization process makes training and rendering faster than previous methods\cite{barron2022mipnerf360, mueller2022instant}. By optimizing the multivariate Gaussians parameterized by position $\mu$, anisotropic covariance $\Sigma = RSS^TR^T$, opacity $\alpha_i$ and color $c_i$ information via posed multi-view images, 3DGS achieves an accurate representation of the scene with accelerated training and rendering times. The color $\mathbf{C}(x')$ and depth $\mathbf{D}(x')$ of each pixel $x'$ in the image can be obtained via point-based volumetric rendering:
\begin{equation} \label{eqn:render}
\begin{split}
        &\mathbf{C} (x') = \sum_{i \in \mathcal{N}} c_i \sigma_i \prod_{j=1}^{i-1}(1- \sigma_j), \,\\
        &\mathbf{D} (x') = \sum_{i \in \mathcal{N}} d_i \sigma_i\prod_{j=1}^{i-1}(1- \sigma_j), 
         \sigma_i = \alpha_i  \mathbf{G'}(x')
\end{split}
\end{equation}
where $\mathcal{N}$ is ordered points overlapping the pixel, and $d_i$ is the distance between 3D Gaussian center and camera center. 
%



\section{Method} \label{sec:method}

Given an unbounded 3D Gaussian Splatting (3DGS) representation~\cite{kerbl3Dgaussians} scene, our objective is to perform text-guided localized 3D scene editing, specifically object insertion or replacement.
We first leverage an off-the-shelf method~\cite{qin2023langsplat} to localize a 3D bounding box as the target editing region. Then, we sample and render azimuth camera viewpoints around the bounding box to feed into our proposed Anchor View Proposal (AVP) module (Sec.~\ref{sec:method-1}), which votes for a view to be used in the 3D lifting step.
Next, we take a user-specified text prompt $\hat{y}$ as guidance to inpaint the anchor view, obtaining an image $\mathbf{x}_{inpaint}$. The object foreground of $\mathbf{x}_{inpaint}$, $\mathbf{x}_{inpaint}^{fore}$, is segmented out by ~\cite{kirillov2023segment} and fed into our coarse-to-fine 3D generation and texture enhancement pipeline (Sec.~\ref{sec:method-2}). In experiments, we assume the generation objects are opaque.
The overview of our pipeline is illustrated in Fig. \ref{fig:pipeline}.

\subsection{Anchor View Proposal} \label{sec:method-1}


Existing multi-view diffusion models~\cite{liu2023zero1to3,ye2023consistent} demonstrate the ability to generate consistent illumination across multi-views given the baked-in lighting in input single-view image.
Inspired by this observation, we aim to identify an anchor viewpoint around the bounding box that includes the strongest illumination cues, such as shadow and highlight. 
To this end, we propose an Anchor View Proposal (AVP) algorithm to reliably select a well-conditioned view from rendering of $N_{anc}$ surrounding viewpoints looking at the center of bounding box. Practically, we define a camera trajectory to render around the object and ensure capture of environmental illumination and no occlusion. The step may need certain amount of manual tuning such as distance to the object and elevation angle if the object in contact with a surface such as a table. The ideal proposed view should contain the most significant difference in illumination between the left-right, top-bottom, or diagonal halves of the rendered images (see evaluation in Sec.~\ref{sec:experiments-qualitative} and Sec.~\ref{sec:experiments-ablation}). This is achieved by rotating each rendered image by a predefined set of Azimuth angles and selecting the rotation that places majority of brighter pixels to the left. For robust lighting estimation from images, we convert rendered RGB images to HSV color space. This design choice is enlightened by the fact that V is the distance along the axis that turns black into non-black color, also known as ``brightness''. Therefore, computation of V-channel image estimates amount of light that lit the surface. In scenarios where the majority of illumination is predominantly cast in the top-bottom direction, our algorithm tends to output an arbitrary view. This behavior aligns with our requirement, as we sample viewpoints in azimuth and any view is considered equally suitable. Pseudocode of AVP algorithm is provided in the supplementary material. 
\subsection{Context-aware Coarse-to-Fine 3D Generation}\label{sec:method-2}
After acquiring the anchor view, we use an off-the-shelf depth-conditioned diffusion model~\cite{zhang2023adding} to inpaint the mask of bounding box projection, according to the generation text prompt $\hat{y}$. The foreground served as input to the next 3D lifting pipeline can be extracted by the segmentation model~\cite{kirillov2023segment}. For efficient and robust 3D generation with contextual illumination awareness, we propose to conduct a coarse but swift image-to-3D generation via 3D-aware diffusion prior, and an illumination-aware texture enhancement step to further realize high-quality 3D lifting result. Note that after removal (if any), we leave the scene points untouched and only optimize new generated points. 


\vspace{-.3cm}

\subsubsection{Coarse Image-to-3D Generation}

Recent efficient multi-view diffusion models~\cite{liu2023zero1to3} were pre-trained on large-scale 3D objects with backed lighting. Given the lighting-heuristic inpainted anchor view, in the coarse step, the multi-view diffusion model can faithfully lift the inpainted image to 3D with multi-view consistency. 


To achieve reliable 3D lifting in 3DGS representation, we adopt compact-based densification and pruning strategy proposed in \cite{chen2023text}. Instead of initializing 3D Gaussians from Point-E~\cite{nichol2022pointe}, we initialize them in a solid sphere, with opacity in a normal distribution proportional to the center distances. This strategy often performs better in our coarse step, since the inpainted conditioning image may not align with the generation modality in Point-E, leading to inconsistent points with the input image and longer training time to converge. Still, some initial points may remain as floaters and are hard to be pruned later. Therefore, we significantly lower initial opacity and colors of 3D Gaussians.


During object optimization by randomly sampled views, we aim to minimize the objective function:
\begin{equation}
    \mathcal{L}=\lambda_{rgb}\mathcal{L}_{rgb} + \lambda_{mask} \mathcal{L}_{mask} + \lambda_{3D-SDS}\mathcal{L}_{3D-SDS},
\end{equation}
where \(\mathcal{L}_{rgb}\) is the mean squared error (MSE) between the foreground anchor view image $\mathbf{x}_{inpaint}^{fore}$ and rendered image at the same viewpoint, \(\mathcal{L}_{mask}\) is MSE loss between ground-truth mask extracted from $\mathbf{x}_{inpaint}^{fore}$ and predicted opacity image $\mathbf{m}_{anchor}$, $\mathcal{L}_{3D-SDS}$ is score distillation loss (Eqn. \ref{eqn:sds}) from 3D-aware diffusion prior~\cite{liu2023zero1to3} for the sample view. The weight of reference view RGB ($\lambda_{rgb}$) increases as training proceeds. We compute $\mathbf{m}_{anchor}$ by volumetric rendering equation:
\begin{equation}
    \mathbf{m}_{anchor} = \sum_{i \in \mathcal{N}} \mathbf{1}\cdot \sigma_i \prod_{j=1}^{i-1}(1- \sigma_j).
\end{equation}
With $\mathcal{L}_{mask}$, we encourage foreground points to be fully opaque and background points to be transparent.

\vspace{-.3cm}

\subsubsection{Illumination-aware Texture Enhancement} \label{sec:method-3}
With the first coarse generation step, the synthesized object has a multi-view consistent shape and appearance, but lacks subtle texture details and high resolution required by photorealism. Therefore, we propose a contextually illumination-aware Texture Enhancement step, which not only enriches geometry and texture details, but also preserves multi-view lighting conditions.



While we observe pre-trained diffusion model with SDS (Eqn.~\ref{eqn:sds}) can greatly enhance texture and geometry details, it does not receive background information from the scene. Inspired by 2D inpainting, we propose Depth-guided Inpainting Score Distillation Sampling (DI-SDS), where intermediate output features of depth-guided ControlNet copy network $\mathcal{C}_{depth}$, projected bounding box mask and masked image are fed into the diffusion model for conditioned generation. Specifically, we first feed colored image latents $x_t$, timestep $t$, text prompt embeddings $y$ and depth image $\mathbf{d}$ into ControlNet layers $\mathcal{C}_{depth}$ to obtain high-level residual block output samples $D$ and middle-level output samples $M$: 
\begin{equation} \label{eqn:1}
    (D, M) = \mathcal{C}_{depth}(x_t;t,y,\mathbf{d}),
\end{equation}
where depth image $\mathbf{d}=1-d$ (i.e., inversed depth), and $d$ is derived from volumetric rendering process (Eqn.~\ref{eqn:render}) and scaled into range $[0, 1]$.

Then, we concatenate bounding box mask $m$ and masked image latents $m_l$ with colored image latents, and feed them additionally into 2D diffusion model $\phi$ with $D,M$:
\begin{equation}
\begin{split} \label{eqn:2}
    &\nabla_\theta \mathcal{L}_{DI-SDS}(\phi, \mathbf{x})=\\
    &\mathbb{E}_{t,\epsilon}[w(t)(\epsilon_\phi (x_t,m,m_l;t,y,D,M)
    -\epsilon)\frac{\partial \mathbf{x}}{\partial \theta}],
\end{split}
\end{equation}
where $\theta$ is 3DGS scene parameters, $\epsilon_\phi(\cdot)$ is predicted sampled noise, $\epsilon$ is a Gaussian noise, $\mathbf{x}$ is rendered RGB image. 

We accommodate classifier-free guidance by computing predicted noise with the following equation: 
\begin{equation} \begin{split}
    &\hat{\epsilon_\phi}(x_t,m,m_l;t,y,\mathbf{d}) = \epsilon_\phi(x_t,m,m_l;t,y,\mathbf{d}) \\&+ s\cdot (\epsilon_\phi(x_t,m,m_l;t,y,\mathbf{d})-\epsilon_\phi(x_t;t)),
\end{split}
\end{equation}
where $s$ is guidance scale specified in Sec.~\ref{sec:experiments-implementation-details}.

The text prompt guiding generation $\hat{y}$ is embedded into view-conditioned text embeddings $y$, for example, ``a standing pineapple, photorealistic, front view'', to prevent Janus problem and provide multi-view information. Since we optimize both object geometry and texture through finetuning, we lower the learning rate in geometry with a focus on smooth texture refinement.
During optimization, we merge Gaussian splats of background color field and generated object splats, and render randomly sampled views following strategy in ~\cite{threestudio2023}. In Texture Enhancement step, we disable $\mathcal{L}_{3D-SDS}$ because it prevents details generated by 2D inpainting diffusion model that does not share the same prior distribution. Therefore, the loss function of this step becomes
\begin{equation} \label{eqn:3}
    \mathcal{L}=\lambda_{rgb} \mathcal{L}_{rgb} + \lambda_{mask} \mathcal{L}_{mask} + \lambda_{DI-SDS} \mathcal{L}_{DI-SDS}.
\end{equation}

Providing flexible controllability for users, this Texture Enhancement step can be balanced by fewer optimization iterations in some cases when simple geometry and texture patterns are sufficiently satisfactory. For example, the monkey statue is expected to have smooth skin in Fig.~\ref{fig:qualitative}.

\begin{figure}[ht]
  \centering
  \includegraphics[width=.7\linewidth]{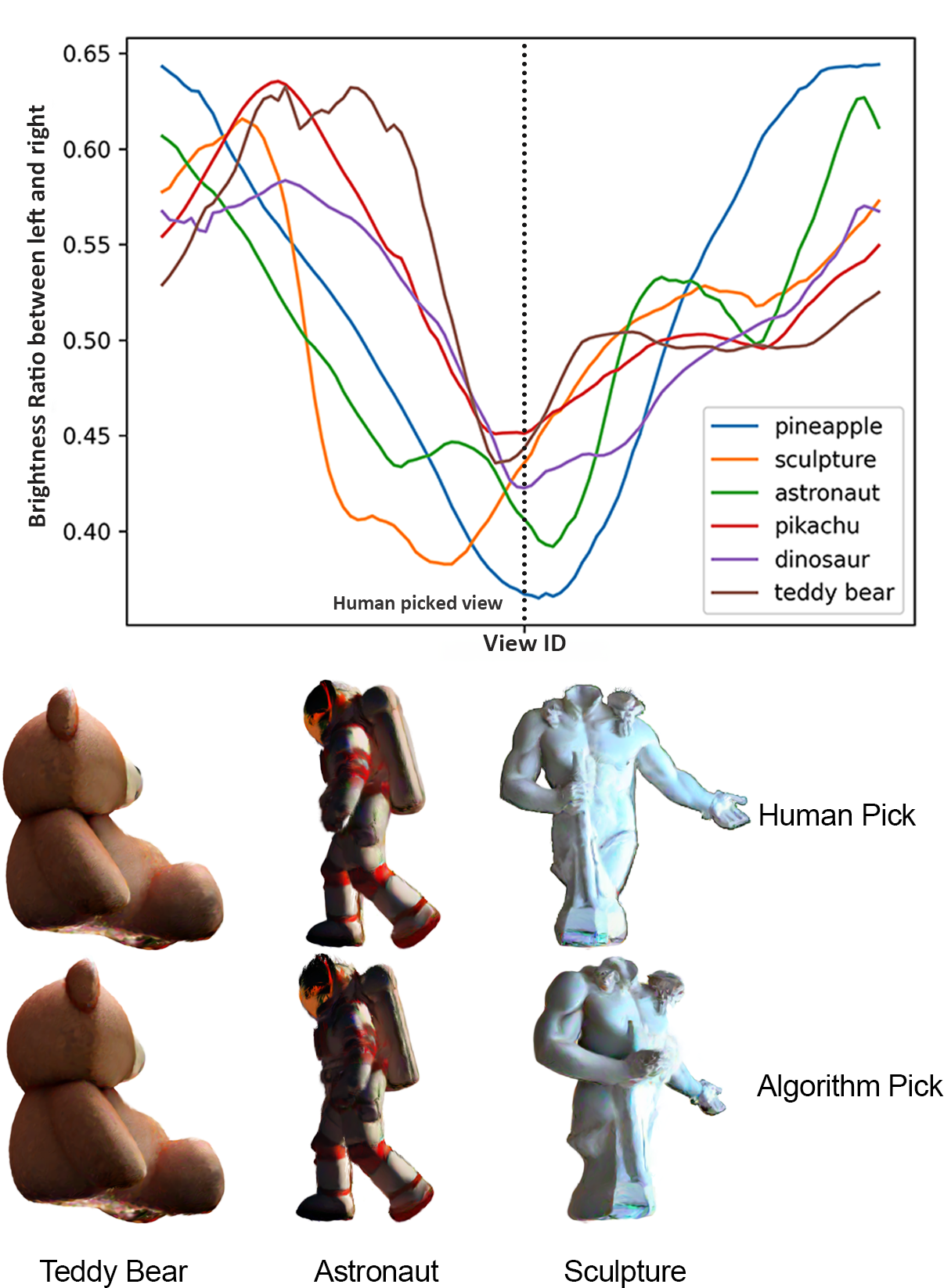}
  \caption{Evaluation on Anchor View Proposal algorithm.
We evaluate the robustness of our Anchor View Proposal (AVP) algorithm by comparing view angle differences between human and algorithm proposals on six objects. The top graph shows the Value-channel (in HSV space) brightness ratio (range: $[0, 1]$) for 100 continuous azimuth views, where a lower value indicates the left side is darker. In this context, AVP selects views with the lowest difference ratio, closely matching human selections. The bottom examples further demonstrate AVP's effectiveness in selecting high contrast views.}
  \label{fig:anchor-view-proposal}
\end{figure}
\begin{figure}[ht]
  \centering
  \includegraphics[width=\linewidth]{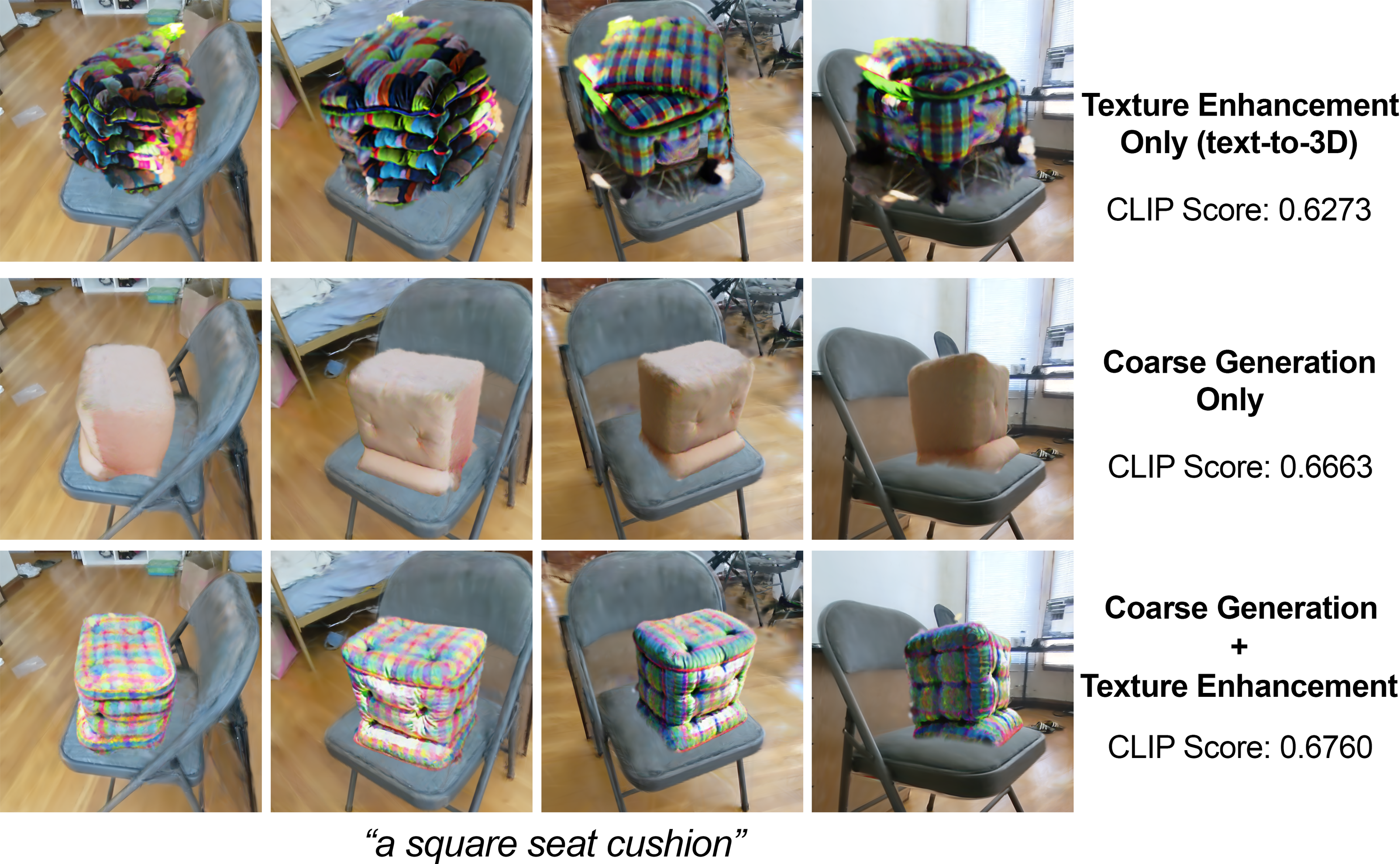}
  \caption{Ablation study on Contextual 3D Lifting.
  We show novel views of localized object editing when we skip coarse step, skip Texture Enhancement step, and apply full pipeline. Coarse step aims to provide initial prior and scene illumination in texture. Texture Enhancement aims to generate more details in geometry and texture.} 
  \label{fig:ablation-step1vsstep2}
\end{figure}
\begin{figure}[ht]
  \centering
  \includegraphics[width=1\linewidth]{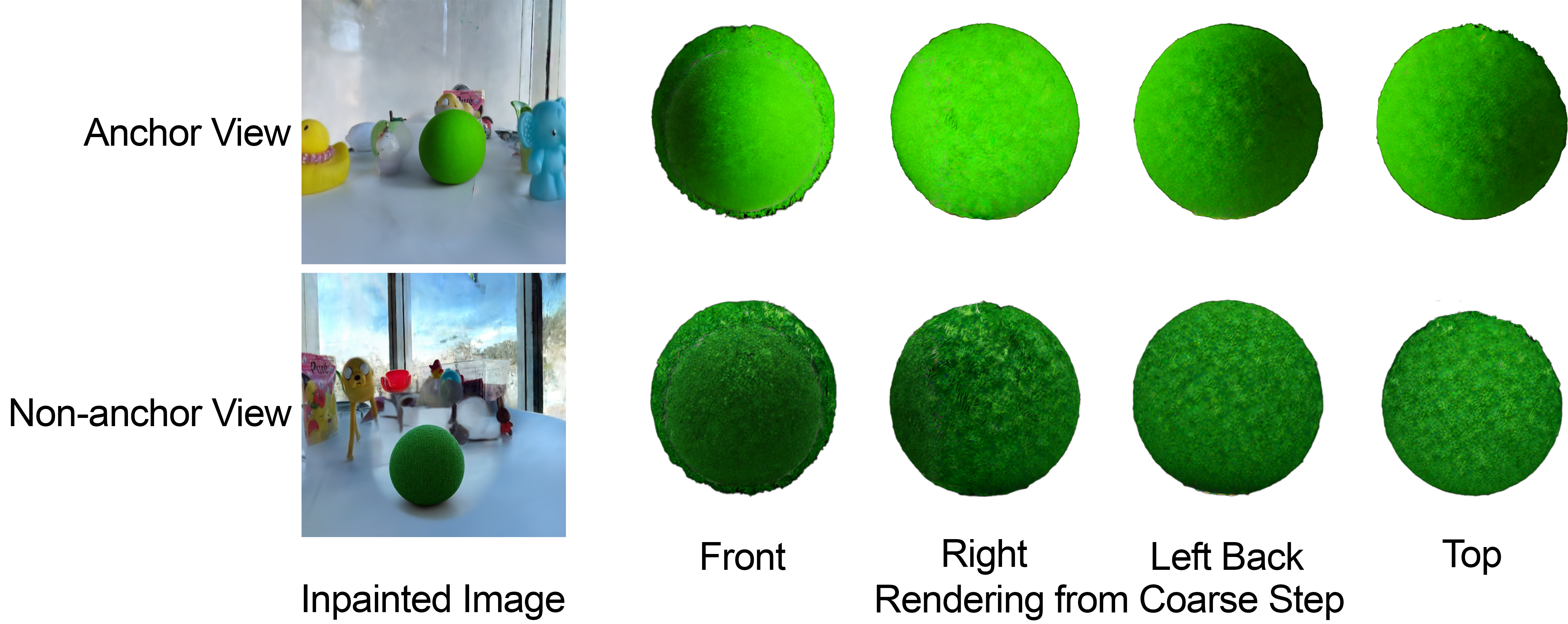}
  \caption{Ablation study on Anchor View Proposal in terms of influence to image-to-3D result. Top row: proposed anchor view. Bottom row: non-anchor view.
  We show inpainted images with the same generator seed in the left column and multi-view rendering of 3D lifted object. Conditioning view that does not reflect scene illumination fails to generate texture that fits into the scene from other viewpoints.} 
  \label{fig:ablation-illumination}
\end{figure}
\begin{figure*}[ht]
  \centering
  \includegraphics[width=0.8\linewidth]{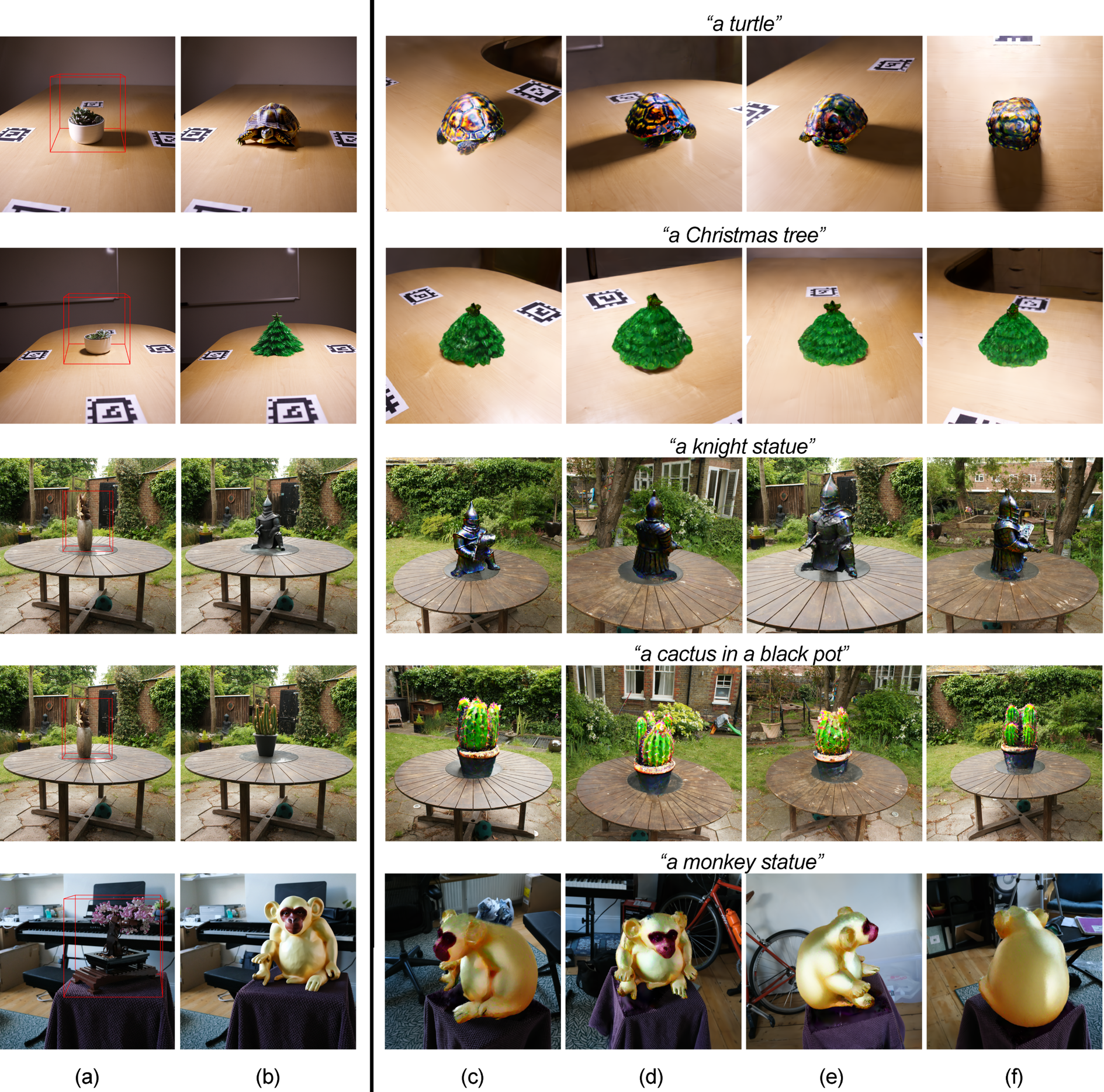}
  \caption{Qualitative Results.
  (a) is input pre-trained 3D Gaussian Splatting scene with bounding box marked in red. (b) is anchor view image with inpainted object according to text prompt on top of each row. (c-f) are multi-view rendering of output scenes. More results are presented in supplementary material.}
  \label{fig:qualitative}
\end{figure*}
\begin{figure*}[h]
  \centering
  \includegraphics[width=\linewidth]{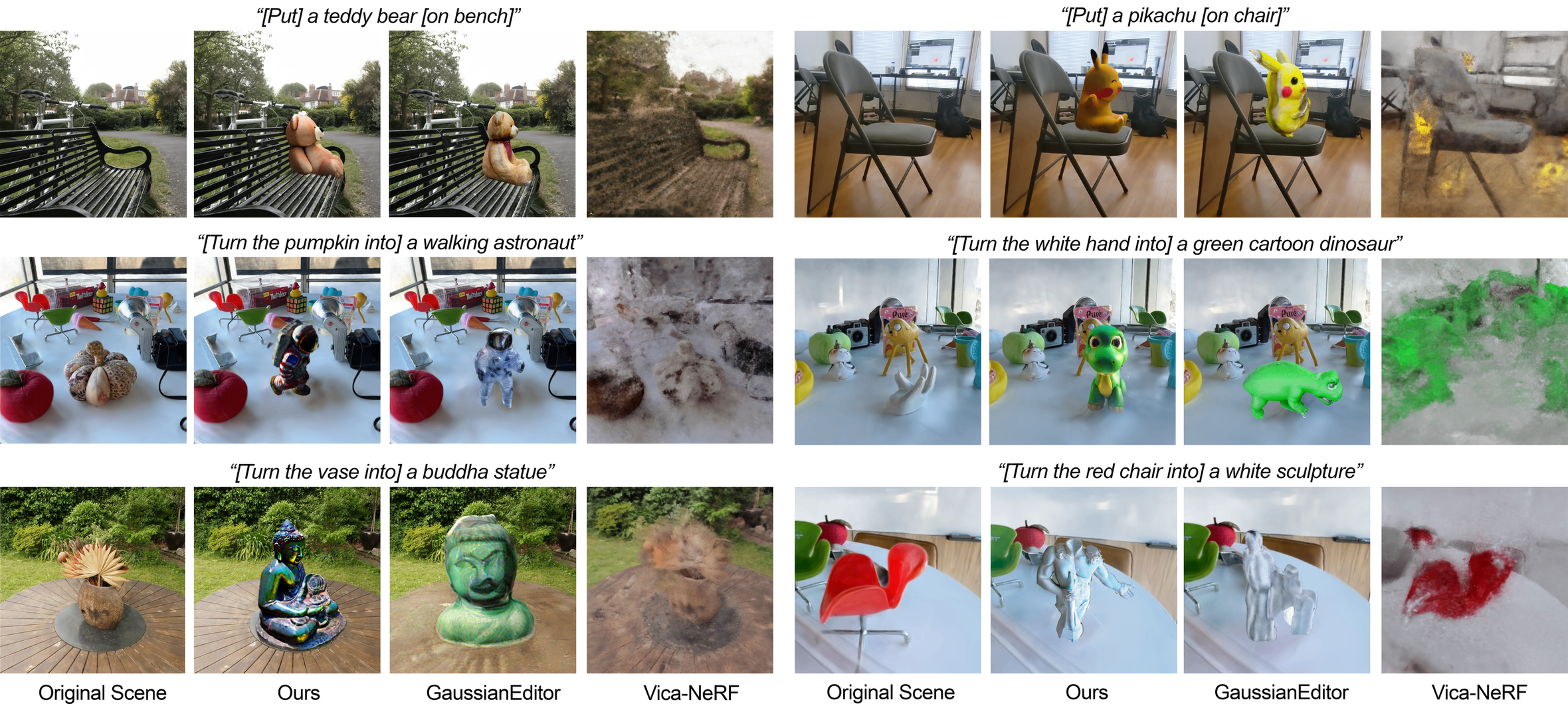}
  \caption{Comparison.
   We show visual results from ours, GaussianEditor \cite{chen2023gaussianeditor} and Vica-NeRF \cite{dong2023vicanerf}, along with a reference image of original scene. Specifically, we provide 1) longer text prompt (with words in square bracket) on top of each row containing localization information to Vica-NeRF; 2) shorter text prompt (without words in square brackets) and pre-determined bounding box to ours and GaussianEditor.}
  \label{fig:comparison}
\end{figure*}

\section{Experiment} \label{sec:experiment}


\subsection{Implementation Details} \label{sec:experiments-implementation-details}
\paragraph{Datasets.}
To evaluate performance in diverse scenes, we use LERF~\cite{lerf2023}, MipNeRF360~\cite{barron2022mipnerf360} and two self-captured datasets where we set up directional light sources. We focus on scenes with challenging illumination that may include multiple objects to test robustness of our pipeline.

\vspace{-.3cm}
\paragraph{Pre-trained Models.} 
We use released Stable Zero123 model in coarse step, ControlNet Depth version~\cite{zhang2023adding} in 2D inpainting and Texture Enhancement step, and Stable Diffusion~\cite{Rombach_2022_CVPR} as 2D diffusion model. Specifically, we leveraged Stable Diffusion Inpainting model by Runway as 2D diffusion model in Texture Enhancement. For view conditioned text embedding, we used Stable Diffusion v1.5 by Runway. Forwarding through ControlNet Depth and Stable Diffusion together provides predicted noise residual. Parameters of all diffusion models are frozen in training.


\subsection{Qualitative Results} \label{sec:experiments-qualitative}

We mix and match text prompt with scenes for generation and render multi-view images for qualitative evaluation in Fig.~\ref{fig:qualitative}. In the figure, we show the rendering of input scene under anchor view with 3D bounding box visualized, and 2D inpainted object under proposed anchor view. Then, we show four multi-view images rendered from output scene. Our result demonstrates detailed texture and faithful respect to scene illumination.
\vspace{-0.3cm}
\paragraph{Evaluation on Anchor View Proposal algorithm.} 

We selected six bounding boxes and their associated objects with strong illumination, then generated multi-view images as input into AVP algorithm. Specifically, we sample 100 object-centered viewpoints along azimuth with every two nearby views are 3.6 degrees apart. Then, we had human manually pick the ``best'' view with the strongest highlight or shadow on left or right half depending on requirements. We graph the offset in viewpoint indices as selected by both the human and our algorithm, and present the three pairs of views with the highest offsets in Fig.~\ref{fig:anchor-view-proposal}. In the plot, we observe that the difference in brightness between left and right half strongly implies the optimal viewpoint to pick. In the case with the highest offset (i.e., the white sculpture), both the human-picked and the algorithm-proposed views are good enough as conditioning viewpoint.


\subsection{Comparison} \label{sec:experiments-comparison}
\paragraph{Qualitative Comparison.} We compare our method against Vica-NeRF~\cite{dong2023vicanerf} and GaussianEditor~\cite{chen2023gaussianeditor} by presenting qualitative result in Fig.~\ref{fig:comparison}. While we additionally provide localization information in text prompt to Vica-NeRF, it is unable to edit the region of interest or preserve background color. Same as ours, GaussianEditor performs segmentation via SAM and 3D lifting from 2D inpainted image. However, GaussianEditor neither includes a mechanism to pick views containing illumination information, nor uses background in multi-view optimization~\cite{long2023wonder3d}. Therefore, it suffers from negligence to the scene illumination such as Pikachu and astronaut cases in Fig.~\ref{fig:comparison}. 

\vspace{-0.3cm}
\paragraph{Quantitative Comparison.} 
We conducted a user study on the results and collected statistics of user's preference. In the questionnaire, we put outputs of different methods in random order and ask participants which one is the most consistent with the text prompt and looks the most realistic within the scene. Statistics show that 0\% users prefer Vica-NeRF, 29.49\% prefer GaussianEditor and 70.51\% prefer ours. 

\subsection{Ablation Study} \label{sec:experiments-ablation}

\paragraph{Effect on Coarse-to-Fine 3D Generation.} In Fig.~\ref{fig:ablation-step1vsstep2}, we show multi-view rendering of a generated square cushion on a chair when we use texture enhancement only, coarse step only, and full pipeline. Without coarse step, the generated object suffers from inconsistency in multi-view and weak fitting to scene illumination. Without Texture Enhancement, we observe consistent geometry, smooth texture, and faithful lifting from anchor view image; however, more details in texture can be added while generation is still aligned with text prompt. With both steps, we observe a significant amount of detail added to both geometry and texture while highlight and shadow cast on the surface still respect the scene. We provide average CLIP score of multi-view rendering following the same way in quantitative comparison (Sec.~\ref{sec:experiments-comparison}).
\vspace{-.5cm}
\paragraph{Effect on Anchor View Proposal.} In Fig.~\ref{fig:ablation-illumination}, we show multi-view rendering of coarse step output when we inpaint the proposed anchor view and another non-anchor view with same inpainting seed and input text prompt ``a green ball''. With the proposed inpainted anchor view, our coarse step and 3D-aware diffusion prior are robust enough to generate multi-views in contextually consistent illumination. The anchor view is more suitable for 3D photorealism, as it implies heuristic illumination information.


\section{Conclusion} \label{sec:conslusion}

In this paper, we proposed an end-to-end pipeline to address text-guided localized editing of 3D scenes, focusing on generation of objects that naturally fit into the scene with illumination awareness. We observed that pre-trained 2D inpainting diffusion model is able to generate objects with texture consistent with scene illumination. Inspired by this, we proposed an Anchor View Proposal algorithm to find a viewpoint, under which we can obtain a conditioning image that contains the most contextual lighting information. We then proposed context-aware coarse-to-fine 3D generation pipeline that faithfully lifts conditioning image into 3D via a coarse step, and then adds details into geometry and texture via a Texture Enhancement step. To resolve the bottleneck that previous SDS does not support optimization with rendering background, we derived DI-SDS from depth inpainting ControlNet to supervise the Texture Enhancement. Experiments demonstrate that our pipeline successfully distills directional illumination from the input scene. Especially, we want to inspire the field to explore derivation and extension to SDS in conditioned generation and editing tasks.

\section*{Acknowledgement}

This research is sponsored by the U.S. Army under contract number W911NF-14-D-0005, Cooperative Agreement number W911NF-20-2-0053 and W911nf-25-2-0040 and by the Office of Naval Research under contract number N00014-21-S-SN03. The views and conclusions contained in this document are those of the authors and should not be interpreted as representing the official policies, either expressed or implied, of the Army, Navy, or the U.S. Government. The U.S. Government is authorized to reproduce and distribute reprints for Government purposes notwithstanding any copyright notation herein.

\newpage
\twocolumn[
    \begin{center}
\LARGE \bfseries
Localized Gaussian Splatting Editing with Contextual Awareness\\[0.5ex]
Supplementary Material
\end{center}
\vspace{1em}
]

\begin{abstract}
    In this supplementary material, we first detail the implementation specifics in Sec.~\ref{sec:implementation-details}, including the pseudocode of our Anchor View Proposal algorithm, the training configurations, and the hardware setup and training speed. Secondly, we supplement additional qualitative results in diverse scenes in Sec.~\ref{sec:more-results}. 
    Thirdly, we present a user study demonstrating the superiority of our method compared to baseline approaches (Sec.~\ref{sec:user_study}). Fourthly, we showcase our graphical user interface in an example generation task (Sec.~\ref{sec:gui}). Lastly, we discuss on the limitations and potential future directions of our research (Sec.~\ref{sec:discussion}). In addition to this document, we have attached several 360-degree videos and a live demo of our GUI usage.
\end{abstract}

\subsection{Implementation Details} \label{sec:implementation-details}
\subsection{Pseudocode of Anchor View Proposal Algorithm}\label{sec:pseudocode_avp}

We provide pseudocode for the proposed Anchor View Proposal (AVP) algorithm in Alg.~\ref{alg:avp}. For the \texttt{addConstrast} function, we compute the minimum and maximum values in the V-channel image and map pixel values from the range [minimum, maximum] to [0, 1]. For the image plane rotation \texttt{R}, we select eight angles, each two views being 45 degrees apart in the 2D image plane.

\subsection{Training Details} \label{sec:training-details}

\paragraph{Resolution.} In the coarse step, we gradually increase the resolution of sampled random views from $64\times 64$ to $128\times 128$, and then to $256\times 256$ at iterations 0, 200, and 300, respectively. For the conditioning view, we increase the resolution from $128\times 128$ to $256\times 256$, and then to $512\times 512$ at the same iteration milestones. We found that the coarse step often converges around 3500 iterations, generating relatively fine detail. During the Texture Enhancement step, the resolution of both random views and the conditioning view is set to $512\times 512$.

\paragraph{Classifier-free Guidance Scale.} We use a guidance scale of 3.0 in the coarse step and 100 for DI-SDS in the Texture Enhancement step.

\begin{center}
\begin{minipage}[t]{1.\linewidth}
\begin{algorithm}[H]
\scriptsize 
\caption{\footnotesize Anchor View Proposal\\
\textbf{Input:} multi-view RGBA images $I$, leftIsBrighter flag $f_l$\\
\textbf{Output:} anchor view index $\hat{i}$}\label{alg:avp}
\begin{algorithmic}
\Function{anchorViewProposal}{$I$, $f_l$}
  \State $V\gets$ getValueChannel(RGBA2HSV($I$)) \Comment{Value Images}
  \State $M\gets$ getAlphaChannel($I$) \Comment{Mask}
  \State $R\gets [1\ldots 8]\times 45$ \Comment{Image Plane Rotations}
  \State $B\gets \mathbf{0}$ \Comment{Image Brightness Balance}
  \ForAll{$v_i$, $m_i$ \textbf{in} $V$, $M$}
    \State $S\gets \mathbf{0}$
    \State $v_i[m_i]\gets$ addContrast($v_i[m_i]$)
    \ForAll{$r$ \textbf{in} $R$}
        \State $g_{i,r}\gets$ rotateImage($v_i$, $r$)
        \State $m_{i,r}\gets$ rotateImage($m_i$, $r$)
        \State $S[r]\gets$ calculateBrightnessBalance($g_{i,r}$, $m_{i,r}$, $f_l$)
    \EndFor
    \State $B[i]\gets$ min($S-1$)
  \EndFor
  \State $B\gets$ abs($B$)
  \State \Return $\hat{i}\gets$ argmin(B)
\EndFunction

\Function{calculateBrightnessBalance}{$v$, $m$, $f_l$}
    \State $h,w\gets$ imageHeight($v$), imageWidth($v$)
    \State $I_l, I_r\gets$ leftHalf($v[m]$), rightHalf($v[m]$)
    \State $b_l, b_r\gets$ mean($I_l$), mean($I_r$) \Comment{Brightness of Masked Half Pixels}
    \If{$f_l$ \textbf{is} \textbf{True}}
        \State \Return $b_l / (b_l+b_r)$
    \Else
        \State \Return $b_r / (b_l+b_r)$
    \EndIf
\EndFunction
\end{algorithmic}
\end{algorithm}
\end{minipage}
\end{center}

\paragraph{Loss Weights.} In the coarse step, the weight of the RGB loss $\lambda_{rgb}$ and the weight of the mask loss $\lambda_{mask}$ both start from 0, increase to 1500 at iteration 1000, and stay at 1000 for the remaining iterations. The weight of the optimization loss by the 3D-aware diffusion prior $\lambda_{3D-SDS}$ remains constant at 1. During the Texture Enhancement step, $\lambda_{rgb}$ and $\lambda_{mask}$ are constantly 1500, and the weight of the DI-SDS loss $\lambda_{DI-SDS}$ is constantly 0.1.

\paragraph{Learning Rates.} The learning rate for the position of Gaussian splats starts from $1\times 10^{-3}$ and decreases to $10^{-5}$ with exponential decay from iteration 0 to 1000 in the coarse step, and from $10^{-4}$ to $10^{-5}$ from iteration 0 to 10000 in the Texture Enhancement step. The learning rate for the scaling of Gaussian splats is $[5\times 10^{-3}, 5\times 10^{-4}]$ in the coarse step and $[5\times 10^{-4}, 5\times 10^{-5}]$ in the Texture Enhancement step, with the same milestones as the position. The rotation, color, and alpha of Gaussian splats are constantly $5\times 10^{-3}, 10^{-2}, 10^{-2}$, respectively, in both steps.

\subsection{Hardware Requirement \& Speed} \label{sec:hardware-requirement-and-speed}

We train our pipeline on a single NVIDIA A40 GPU. The VRAM usage is approximately 20GB for the coarse step and 43GB for the Texture Enhancement step. Each iteration takes around 0.5 seconds for the coarse step and 7.8 seconds for the Texture Enhancement step, as per the settings detailed in Sec.~\ref{sec:training-details}. The total number of iterations depends on complexity in geometry and texture. Typically, we found the coarse step converges after 6500 iterations. While texture enhancement step highly depends on user's preference, we usually stops this step after 2000 iterations incremented from the coarse step. We sample random views with a batch size of 6 for the coarse step and 4 for the Texture Enhancement step. We discovered that increasing the batch size significantly reduces color deviation in the Texture Enhancement step.

\subsection{More Results} \label{sec:more-results}

We provide more qualitative results in Fig.~\ref{fig:more-results-1}, ~\ref{fig:more-results-2} and ~\ref{fig:more-results-3}. In the figures, we show the anchor view images of the original scenes with 3D bounding box for editing, inpainted anchor view images, text prompts for generation, and multi-view rendering results from our pipeline. In addition to LERF~\cite{lerf2023}, MipNeRF360~\cite{barron2022mipnerf360} and our self-captured datasets, we use a commercially purchased synthetic scene (the first case in Fig.~\ref{fig:more-results-1} and the third case in Fig.~\ref{fig:more-results-3}). For the synthetic scene, we rendered RGB images in $800\times 800$ resolution under 332 virtual camera views ($48.5$-degree field of view) around the scene to densely supervise 3DGS training. 
For our self-captured datasets, we use a Canon EOS-1D camera to capture around 400 photos of each scene and send the raw colored images to COLMAP for calibration.


\subsection{User Study} \label{sec:user_study}

\begin{figure} 
    \begin{center}
        \includegraphics[width=.7\linewidth]{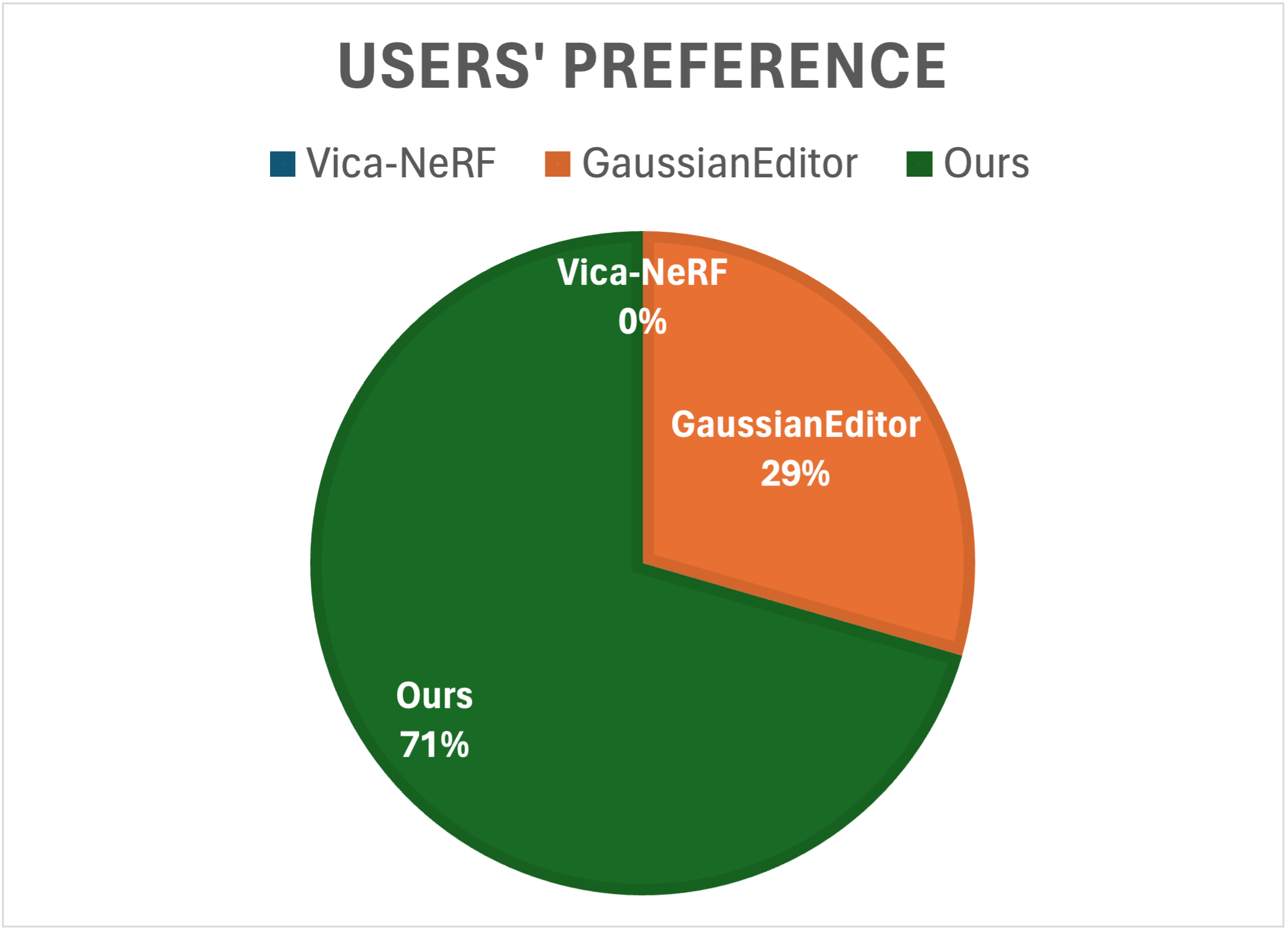}
    \end{center}
    \caption{User Study statistics}
    \label{fig:user-study}
\end{figure}

To further substantiate our claim that our method generates more photorealistic results than state-of-the-art methods -- GaussianEditor~\cite{chen2023gaussianeditor} and Vica-NeRF~\cite{dong2023vicanerf}, we conducted an anonymous user study. The statistics of users' preferences are provided in Fig.~\ref{fig:user-study}.

We created a questionnaire featuring output examples from different methods shown in Fig.5 of the main paper. The order of the methods was randomized to mitigate potential bias. Participants were asked to choose the result that best aligned with the input text prompt and appeared the most photorealistic in each question comparing the three methods.

We also requested participants to share their age group (options were ```below 18'', ``18-25'', ``26-35'', ``36-45'', ``46+'') and their familiarity with the fields of computer vision (CV) or computer graphics (CG) (options were ``Yes'', ``No'', ``Maybe''). All 26 questionnaires were collected anonymously.

According to the results, no users preferred Vica-NeRF, 29.49\% preferred GaussianEditor, and 70.51\% preferred our method. Among the participants, 19.2\% fell into the age group of 18-25, 76.9\% into 26-35, and 3.8\% did not provide their age. As for familiarity with CV or CG, 61.5\% claimed familiarity, 26.9\% claimed no familiarity, 7.7\% were uncertain (``Maybe''), and 3.8\% did not provide their familiarity level.

\subsection{Graphical User Interface} \label{sec:gui}

We also package our pipeline into an application with graphical user interface as an extra contribution. An example of the interface can be found in Fig.~\ref{fig:gui}. The system is built upon Viser~\cite{viser}, an open-source interactive 3D visualization library. The system supports original GSGen~\cite{chen2023text} and 3D Gaussian Splatting~\cite{kerbl3Dgaussians} rendering and live progress check of training. We also leave APIs easy to change in a separate configuration file. This allows users and researchers to conduct future experiments in both text-to-3D and image-to-3D generative tasks.

\subsection{Discussion} \label{sec:discussion}

\subsection{Limitations}

Our current method guarantees photorealism between the generated object and input scene, but it does not ensure physically accurate lighting. In scenes with complex lighting, a single conditioning anchor view often provides insufficient lighting information. Additionally, current text-guided generation schemes do not support shadow editing, which requires disentanglement and editing of background. Therefore, it is necessary to involve conventional inverse rendering and ray tracing similar to~\cite{Zeng_2023,lyu2022neuralradiancetransferfields,srinivasan2020nervneuralreflectancevisibility,Zhang_2021,verbin2021refnerfstructuredviewdependentappearance,liang2024gsir3dgaussiansplatting}. Also, our pipeline may still require manual tuning in AVP step, where the collection of renderings for anchor view proposal needs to avoid occlusion and surface intersection. When the directional illumination is cast from top to bottom, the manual set elevation angle is crucial. It will be intuitive and straightforward to extend Azimuth rotation to further sample also elevation rotation, but this will also require manual tuning.

\subsection{Future Works}

Our current task involves object-centered replacement and editing. This process requires the editing region to be enclosed in a bounding box, separate from the other point clouds in the scene. As a result, detailed editing that interacts with existing scene objects is not yet possible. In such cases, we need a more precise segmentation and generation method to optimize both the generated point cloud and the scene point cloud, to ensure multi-view consistency.

\begin{figure*}
    \centering
    \includegraphics[width=.78\linewidth]{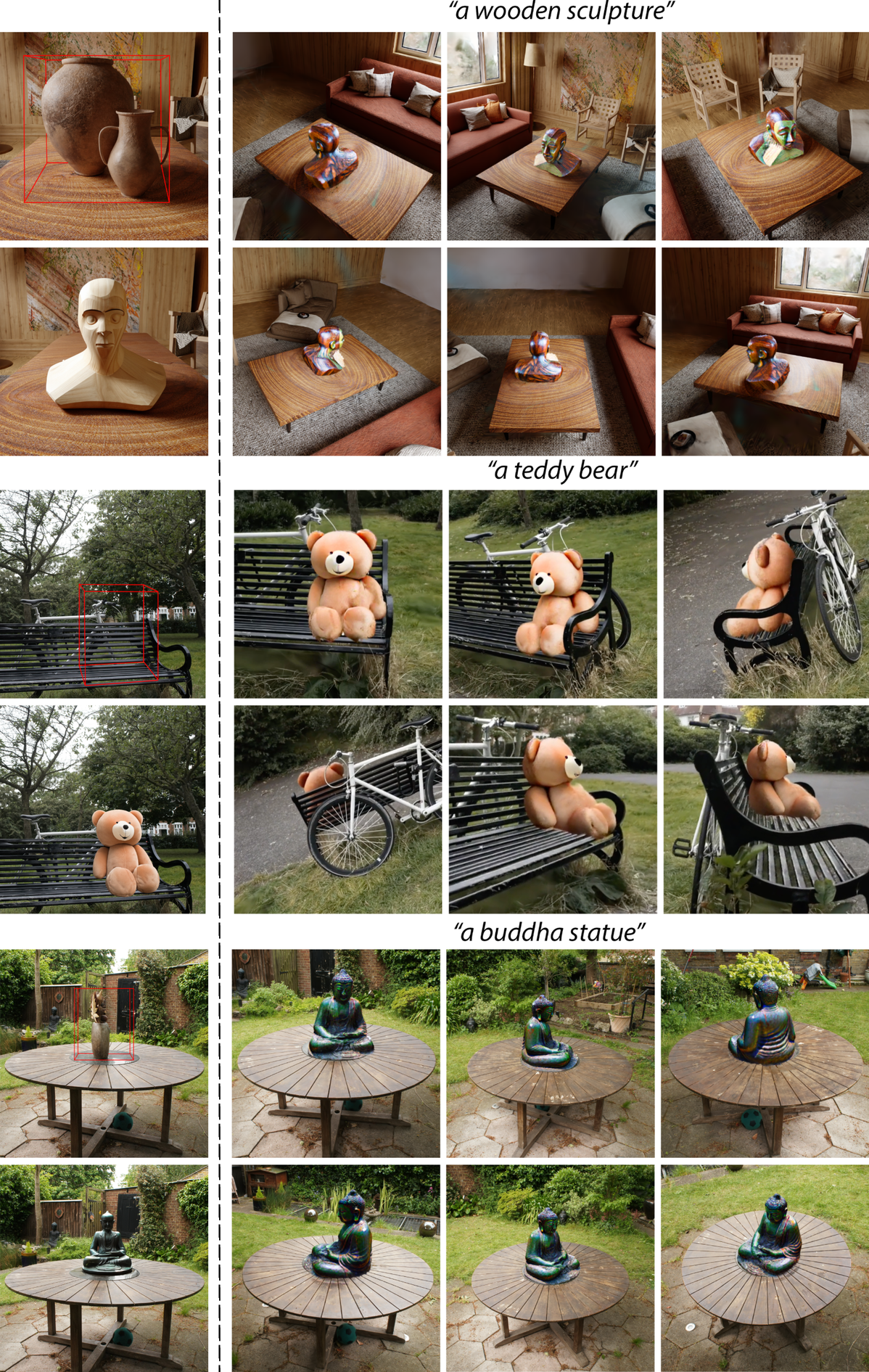}
    \caption{More Results 1 of 3. We show original and inpainted anchor view in the first column, and multi-view rendering results of our method in the right three columns.}
    \label{fig:more-results-1}
\end{figure*}

\begin{figure*}
    \centering
    \includegraphics[width=.78\linewidth]{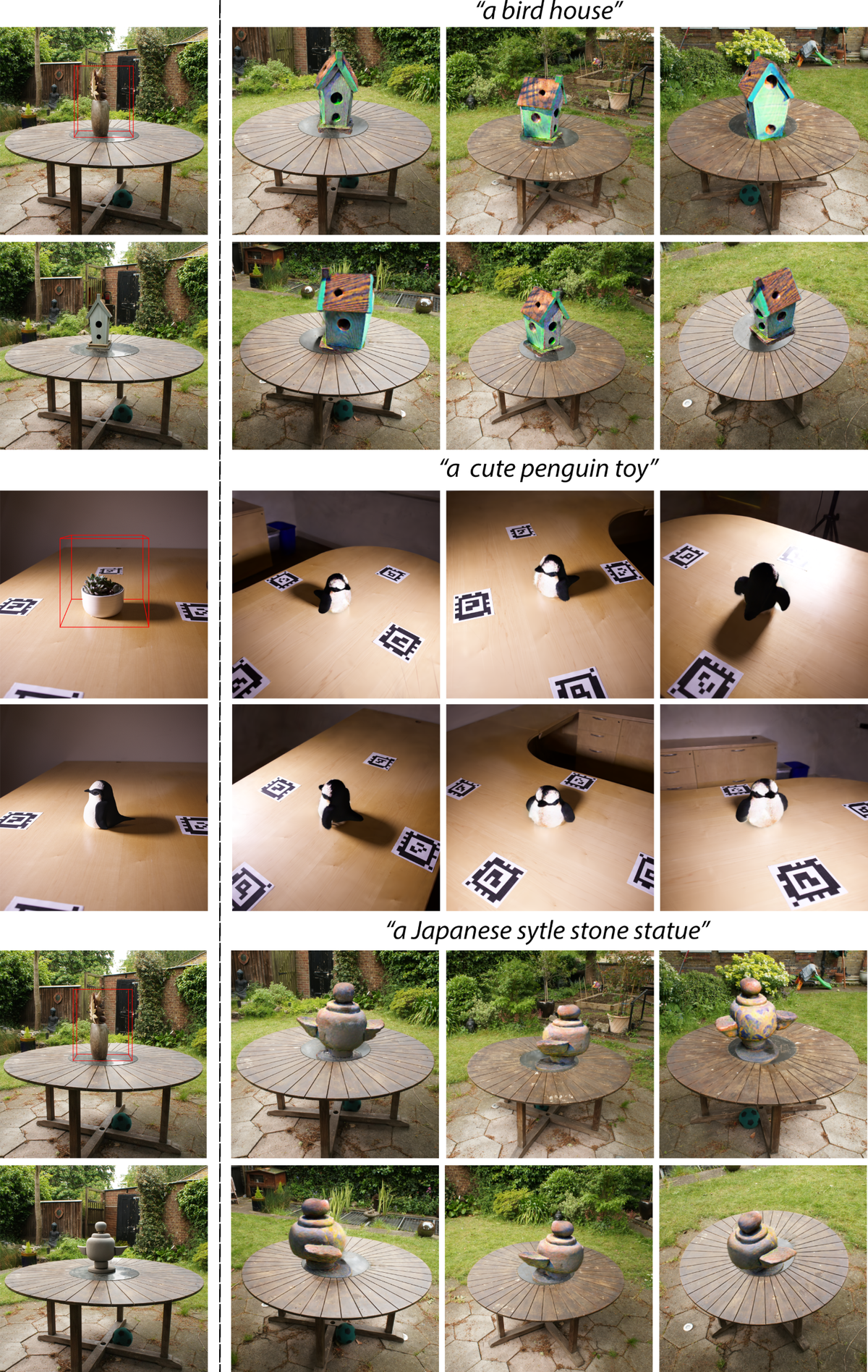}
    \caption{More Results 2 of 3. We show original and inpainted anchor view in the first column, and multi-view rendering results of our method in the right three columns.}
    \label{fig:more-results-2}
\end{figure*}

\begin{figure*}
    \centering
    \includegraphics[width=.78\linewidth]{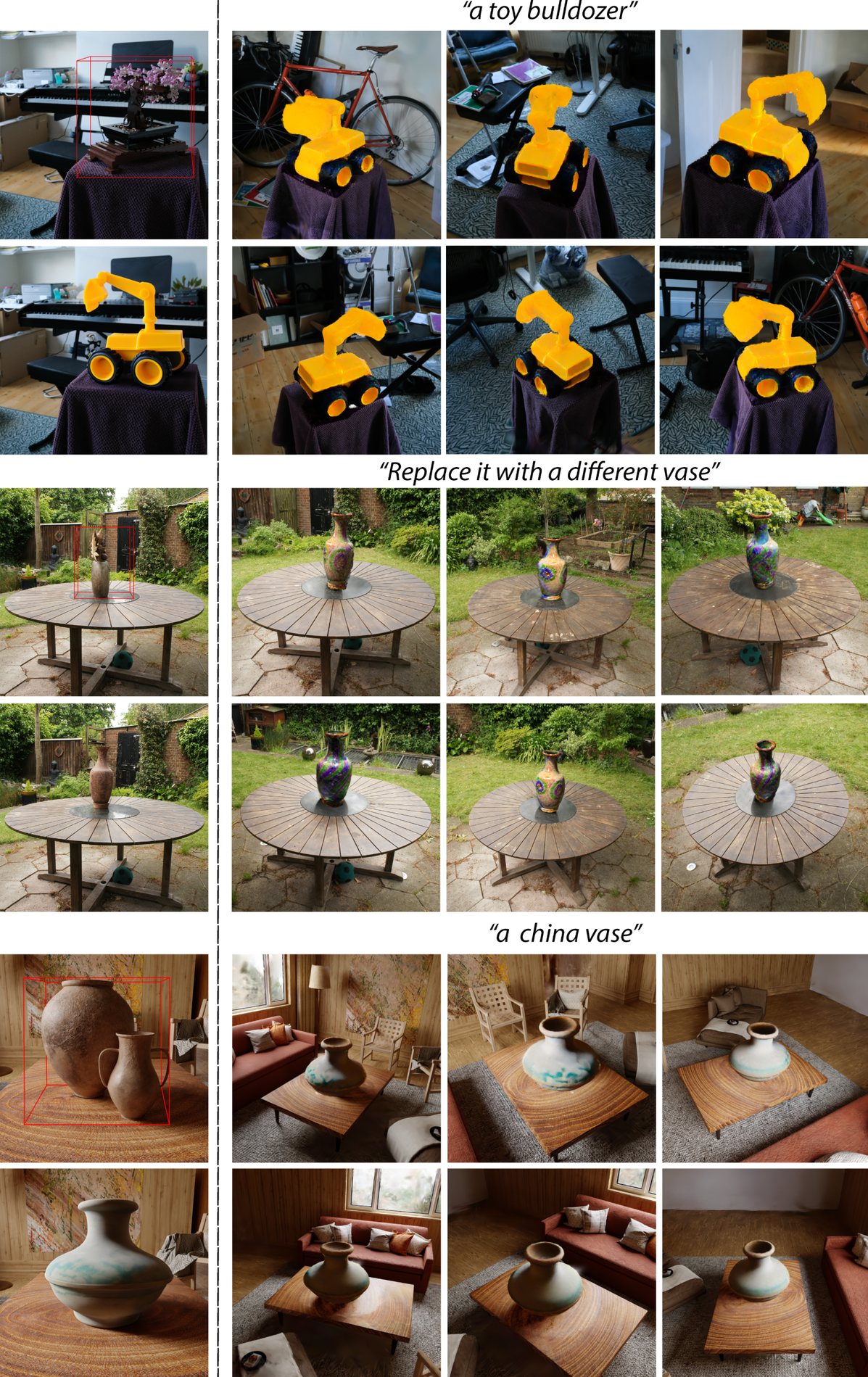}
    \caption{More Results 3 of 3. We show original and inpainted anchor view in the first column, and multi-view rendering results of our method in the right three columns.}
    \label{fig:more-results-3}
\end{figure*}

\begin{figure*}
    \centering
    \includegraphics[width=.9\linewidth]{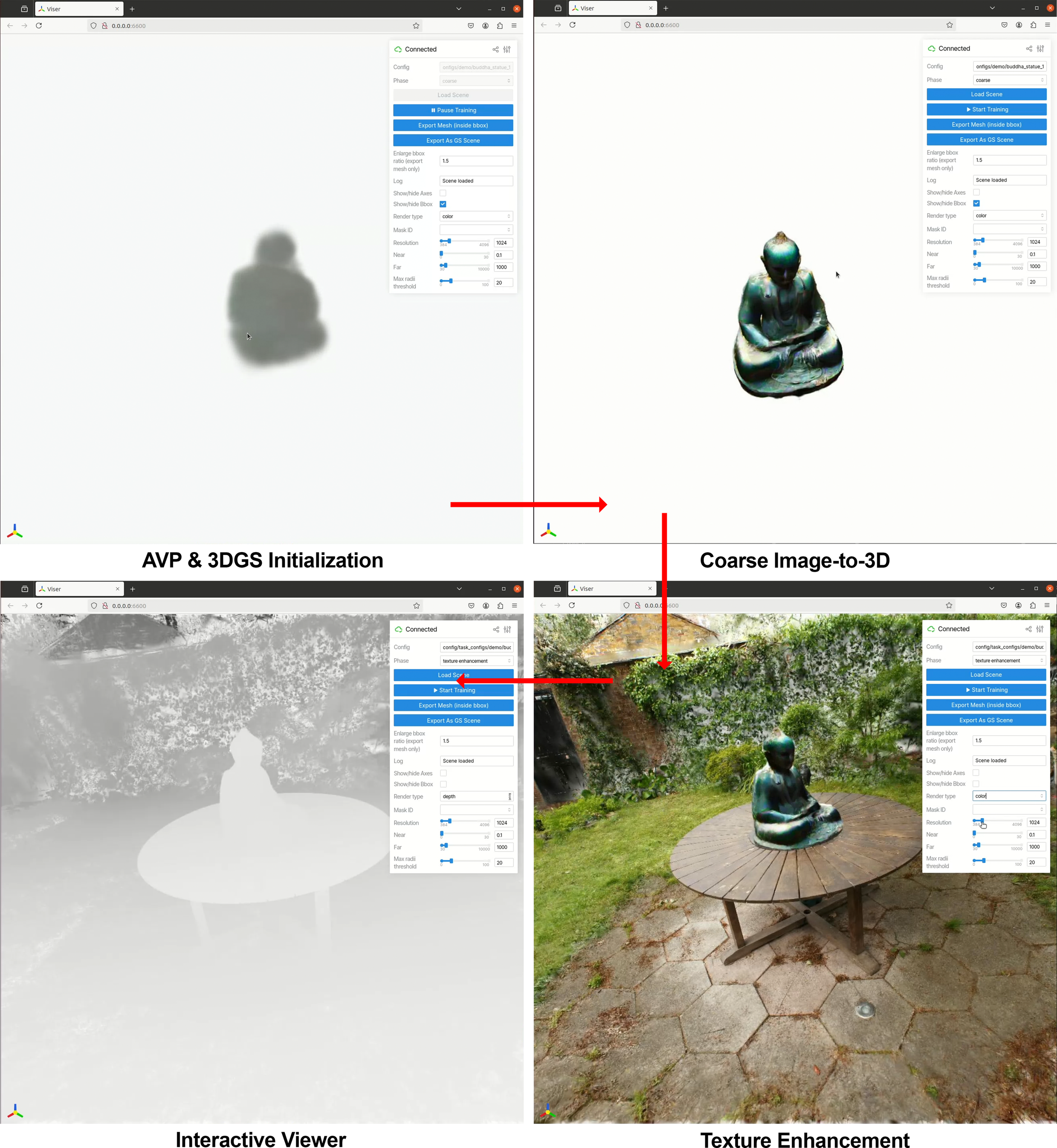}
    \caption{Graphical User Interface of our pipeline. We develop a GUI that includes Anchor View Proposal, coarse image-to-3D generation and texture enhancement steps. The interactive viewer is especially useful to visualize training procedure and results. In the lower left figure, we show rendering of depth images of the same scene besides rendering of RGBA frames.}
    \label{fig:gui}
\end{figure*}


{\small
\bibliographystyle{ieee_fullname}
\bibliography{egbib}
}

\end{document}